\documentclass[sigconf]{acmart}

\usepackage{caption}
\usepackage{amsmath}
\usepackage{amsthm}
\usepackage{booktabs}
\usepackage{algorithm}
\usepackage{algorithmicx}
\usepackage{subfigure}
\usepackage{multirow}
\usepackage{bm}
\usepackage{tablefootnote}
\usepackage{hyperref}
\usepackage{algpseudocode}
\usepackage{amsfonts}
\usepackage{amsmath}
\usepackage{booktabs}
\AtBeginDocument{
  \providecommand\BibTeX{{
    \normalfont B\kern-0.5em{\scshape i\kern-0.25em b}\kern-0.8em\TeX}}}

\copyrightyear{2020} 
\acmYear{2020} 
\setcopyright{acmcopyright}\acmConference[SIGIR '20]{Proceedings of the 43rd International ACM SIGIR Conference on Research and Development in Information Retrieval}{July 25--30, 2020}{Virtual Event, China}
\acmBooktitle{Proceedings of the 43rd International ACM SIGIR Conference on Research and Development in Information Retrieval (SIGIR '20), July 25--30, 2020, Virtual Event, China}
\acmPrice{15.00}
\acmDOI{10.1145/3397271.3401437}
\acmISBN{978-1-4503-8016-4/20/07}

\begin{document}
\fancyhead{}

\title{Network On Network for Tabular Data Classification in Real-world Applications}

\author{Yuanfei Luo}
\email{luoyuanfei@4paradigm.com}
\affiliation{
	\institution{4Paradigm Inc., Beijing, China}
}
\author{Hao Zhou}
\email{zhouhao@4paradigm.com}
\affiliation{
	\institution{4Paradigm Inc., Beijing, China}
}
\author{Wei-Wei Tu}
\email{tuweiwei@4paradigm.com}
\affiliation{
	\institution{4Paradigm Inc., Beijing, China}
}
\author{Yuqiang Chen}
\email{chenyuqiang@4paradigm.com}
\affiliation{
	\institution{4Paradigm Inc., Beijing, China}
}
\author{Wenyuan Dai}
\email{daiwenyuan@4paradigm.com}
\affiliation{
	\institution{4Paradigm Inc., Beijing, China}
}
\author{Qiang Yang}
\email{qyang@cse.ust.hk}
\affiliation{
	\institution{Hong Kong University of Science and Technology, Hong Kong}
}
\renewcommand{\shortauthors}{Luo et al.}

\begin{abstract}
Tabular data is the most common data format adopted by our customers ranging from retail, finance to E-commerce, and tabular data classification plays an essential role to their businesses.
In this paper, we present \textit{Network On Network} (NON), a practical tabular data classification model based on deep neural network to provide accurate predictions.
Various deep methods have been proposed and promising progress has been made.
However, most of them use operations like neural network and factorization machines to fuse the embeddings of different features directly, and linearly combine the outputs of those operations to get the final prediction.
As a result, the intra-field information and the non-linear interactions between those operations (e.g. neural network and factorization machines) are ignored.
Intra-field information is the information that features inside each field belong to the same field.
NON is proposed to take full advantage of intra-field information and non-linear interactions.
It consists of three components: field-wise network at the bottom to capture the intra-field information, across field network in the middle to choose suitable operations data-drivenly, and operation fusion network on the top to fuse outputs of the chosen operations deeply.
Extensive experiments on six real-world datasets demonstrate NON can outperform the state-of-the-art models significantly.
Furthermore, both qualitative and quantitative study of the features in the embedding space show NON can capture intra-field information effectively.
\end{abstract}



\maketitle

\section{Introduction}\label{sec:intro}
\begin{figure}
	\centering 
	\includegraphics[width=0.48\textwidth]{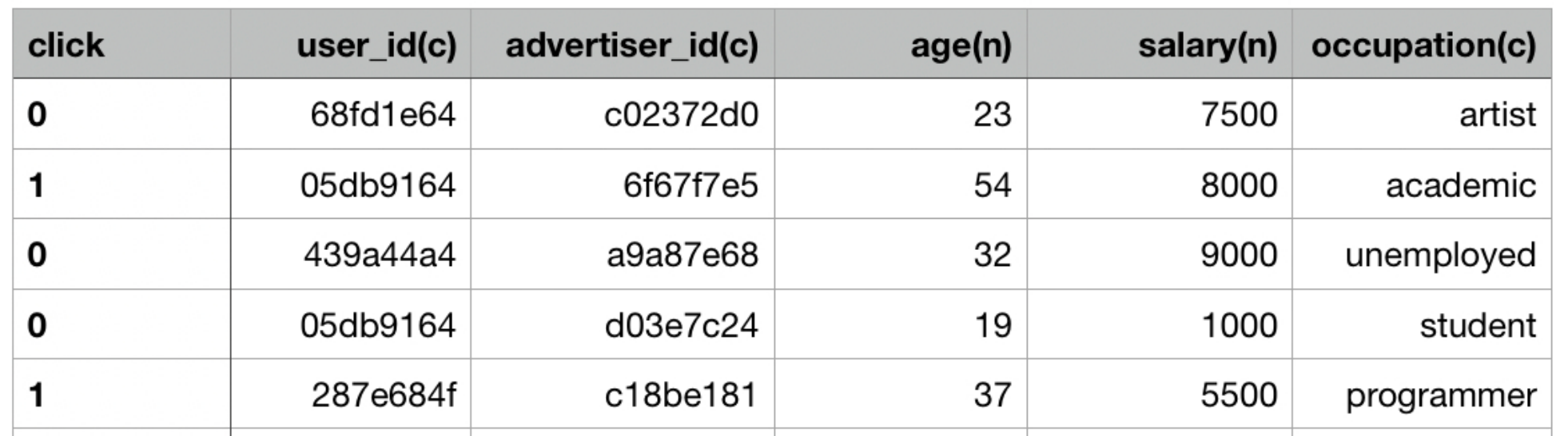}
	\caption{An example of tabular data in online advertising, `click' is 1 if user (`user\_id') click an advertiser's ad (`advertiser\_id'). Letters `n' and `c' embraced in the parentheses indicate `numerical' and `categorical' features respectively.}\label{fig:tabular_data}
\end{figure}

Tabular data is widely used in many real-world applications, such as online advertising~\cite{evans2009online,zeff1999advertising}
recommendation~\cite{bobadilla2013recommender}, fraud detection~\cite{bolton2002statistical,wang2010comprehensive}, medical treatment~\cite{kononenko2001machine}, which is also the scenario of our customers ranging from retail, finance to E-commence.
In tabular data, each row corresponds to an instance and each column corresponds to a field, and tabular data classification is to classify the instance to one or more classes according to its fields. 
Classification performance is crucial to the real-world businesses. As discussed in the paper \cite{cheng2016wide} of YouTube and collected feedbacks from our customers, improve the classification performance can bring remarkable extra revenue to their businesses.
An example of tabular data is shown in Figure \ref{fig:tabular_data}.
Tabular data is often a mixture of numerical fields and categorical fields, and those categorical fields are usually high-dimensional, e.g. in the online advertising, the categorical field `advertiser\_id' may contain millions of distinct advertiser IDs, which makes the classification problem rather challenging.

With strong representation and generalization abilities, deep learning based methods have been widely studied for tabular data classification in recent years and achieve considerable success \cite{cheng2016wide,guo2017deepfm,song2018autoint}.
In Wide \& Deep~\cite{cheng2016wide}, DNN accompanied by a linear model (e.g. logistic regression) is proposed, where the inputs of linear part are handcrafted features.
Deep factorization machine (DeepFM)~\cite{guo2017deepfm} uses FM~\cite{rendle2010factorization}  to replace the linear part of Wide \& Deep to reduce the dependence on feature engineering.
In extreme deep factorization machine (xDeepFM)~\cite{lian2018xdeepfm}, a novel compressed interaction network is proposed to learn feature interactions explicitly, and it is used to replace the linear part of Wide \& Deep. 
AutoInt~\cite{song2018autoint} also follows the structure of Wide \& Deep, and the linear part is replaced by a multi-head self-attentive neural network with residual connections. In short, most of the above models fall into the design paradigm of 1) projecting the input categorical features in each field into low-dimensional embeddings; 2) using multiple operations like DNN or FM to directly fuse the embeddings of different fields; 3) linearly combine the output of each operation to get the final prediction.
While these models have many advantages, three problems remain to be solved.

First, the information that features inside each field belong to the same field are not fully captured, since we fuse all the embeddings of different fields directly without considering this information explicitly. Here we denote this information as \textit{intra-field information}. For features inside each field, their intrinsic property is that they all belong to the same field. Consider online advertising scenario as an example, the fields `advertiser\_id' and `user\_id' represent the ID of advertisers and users respectively. Advertisers are usually companies who want to make their products known by more people and users are those who surf the internet. The information that a specific ID is an advertiser or a user may make classification more accurate. Besides, fields have their own meanings, just like `advertiser\_id" and `user\_id' represent the set of advertisers and users respectively,  regardless of what specific IDs they contain.
Second, most of the existing methods use a predefined combination of operations regardless of data, as summarized in Table \ref{tab:proposed_ops}.
In practice, the predefined combination of operations may not be suitable for all the data and we should choose different operations according to our data for better performance.
Third, non-linear interactions between the operations (e.g. neural network and FM) are ignored by the aforementioned methods. As shown in Table \ref{tab:proposed_ops}, the outputs of different operations are first concatenated, and then weighted sum to get the final prediction.
The non-linear interactions are overlooked because of the inherent linearity of weighted sum.

\begin{table}
	\caption{Operations predefined by existing methods and their way to get the final predictions, where $\bm{h}_{\textup{i}}$ denotes the output vector of operation $i$.}
	\label{tab:proposed_ops}

	\centering
	\renewcommand\tabcolsep{2pt}
	\begin{tabular}{lrr}
		\toprule 
		Method              & Operations &  Prediction\\ 
		\midrule 
		Wide \& Deep \cite{cheng2016wide}    & Linear \& DNN  
		&  $\sigma \left( \bm{w}^{T}\left[ \bm{h}_{\textup{lin}}, \bm{h}_{\textup{dnn}} \right] \right)$ \\
		DeepFM \cite{guo2017deepfm}             & FM \& DNN
		&  $\sigma \left( \bm{w}^{T}\left[ \bm{h}_{\textup{fm}}, \bm{h}_{\textup{dnn}} \right] \right)$ \\
		xDeepFM \cite{lian2018xdeepfm}            & CIN \& Linear \& DNN
		&  $\sigma \left( \bm{w}^{T}\left[ \bm{h}_{\textup{cin}}, \bm{h}_{\textup{lin}}, \bm{h}_{\textup{dnn}} \right] \right)$ \\
		AutoInt \cite{song2018autoint}             & self-attention \& DNN 
		&  $\sigma \left( \bm{w}^{T}\left[ \bm{h}_{\textup{self}}, \bm{h}_{\textup{dnn}} \right] \right)$ \\
		\bottomrule 
	\end{tabular}
\end{table}

To alleviate the above issues and provide better service to our customers, we propose \textit{Network On Network} (NON) for tabular data classification.
NON consists of three parts: field-wise network at the bottom,  across field network in the middle and operation fusion network on the top.
Field-wise network employ a unique DNN for each field to capture the intra-field information. 
Unlike previous works use predefined operations, across field network treats operations like LR as optional, and explore various human-designed operations to choose the most suitable operations for the input data.
Finally the operation fusion network on the top utilize DNN to fuse outputs of the chosen operations deeply.
While the architecture powers the model's expressiveness, it becomes deeper and harder to train. Inspired by GoogLeNet~\cite{szegedy2015going}, we introduce a new training technique to make the training process easier by adding auxiliary classifier to each layer of DNN.
To summarize, this paper has the following contributions:
\begin{itemize}
    \item We propose field-wise network to capture intra-field information. Empirical studies show field-wise network can capture intra-field information and improve generalization.
    \item Based on field-wise network, we further propose a novel architecture named Network On Network (NON), along with an improved training technique to make NON easier to train. 
    \item Extensive experiments on several real-world datasets show NON can outperform state-of-the-art models significantly and consistently.
\end{itemize}

The remaining of this paper is organized as follows: related works are shown in Section \ref{sec:related_work}, and NON is introduced in Section \ref{sec:proposedmethod}. Section \ref{sec:exp} shows the experimental results and Section \ref{sec:conclusion} concludes the paper.

\section{Related Work}\label{sec:related_work}
\begin{figure}
	\includegraphics[width=0.48\textwidth]{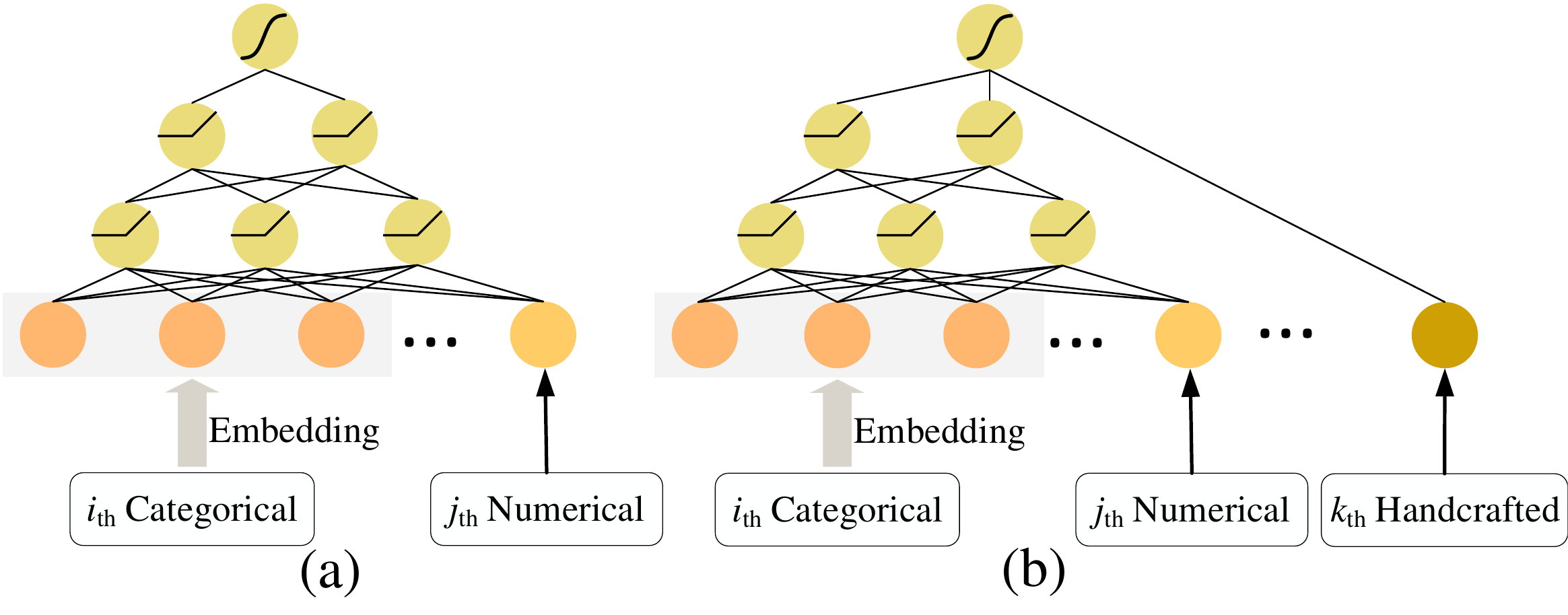}
	\caption{(a) The structure of vanilla DNN for tabular data classification. (b) The structure of Wide \& Deep.} 
	\label{fig:related}
\end{figure}

\subsection{Tabular data classification}
Tabular data is one of the most common data format used in real-world businesses~\cite{yuanfei2019autocross}. Data in tabular is structured into rows, where each row corresponds to an instance and each column corresponds to a field (an attribute), and tabular data classification is to classify the instance to one or more classes according to its fields. 
Tabular data is often a mixture of numerical fields and categorical fields, and the categorical fields are usually high-dimensional. 
Meanwhile, those categorical fields are vital to model personalization and accuracy.
Figure \ref{fig:tabular_data} gives an example of tabular data in online advertising. For categorical fields like `user\_id', it may contain millions (even billions) of distinct user IDs, and features based on this field can reflect users' personal preference.

Tree methods including random forests \cite{breiman2001random} and gradient boosting machine \cite{gbm_2000,chen2016xgboost} usually perform well on tabular data with numerical fields, but they are not friendly to high-dimensional tabular data. On the one hand, for each node of the tree, tree methods need to enumerate all the features for all the fields, which is inefficient for high-dimensional categorical fields. On the other hand, the gain by splitting on a categorical feature will be very small because of rareness of categorical features.
Tabular data with high-dimensional categorical fields is a more general scenario in real-world applications, thus tree methods are not included in the baselines.

\subsection{Shallow methods}
In industrial scenario, logistic regression (LR) was one of the most popular methods for large-scale sparse tabular data classification~\cite{cheng2010personalized,mcmahan2013ad}. But it lacks the ability to learn feature interactions because of its linearity, and lots of feature engineering need to be done.
Factorization Machines (FMs)~\cite{rendle2010factorization} and field-aware factorization machines (FFMs)~\cite{juan2016field} are proposed to embed the sparse input features into low-dimensional dense vectors and use the inner product of the vectors to explicitly learn the 2nd-order feature interactions.
Due to the shallow structures of FMs and FFMs, their representative ability is also limited~\cite{guo2017deepfm}.

\subsection{Deep methods}
Benefits from the embedding vectors and nonlinear activation functions, DNN can implicitly learn the high-order feature interactions. 
The structure of vanilla DNN for tabular data classification is shown in Figure \ref{fig:related}(a). The embedding function in Figure \ref{fig:related}(a) has been widely used to transform the categorical features into low-dimensional dense vectors. 
Generally, the sparse input of tabular data classification can be formulated as 
\begin{equation*}
	\bm{x} = [x_1, x_2,..., x_m] ,
\end{equation*}
where $m$ is the total number of fields, $x_i \in \mathbb{R}^{n_{i}}$ is a one-hot vector for a categorical field with $n_{i}$ features and $x_i \in \mathbb{R}$ is vector with only one value for a numerical field. 
Since the one-hot vector is very sparse and high-dimensional, one can use embedding function to transform it into a low-dimensional vector of real values using $\bm{e_i} = \mathbf{W}_i x_i$,
where $\mathbf{W}_i \in \mathbb{R}^{d \times n_i}$ is the embedding table of $i$-th field and $d$ is the embedding dimension. 
One can also transform the numerical feature field into the same low-dimensional space by $ \bm{e_j} = \mathbf{V}_j x_j$,
where $\mathbf{v}_j\in \mathbb{R}^{d}$ and $x_j \in \mathbb{R}$. 

Product-based Neural Network~\cite{pnn_tois2018} and its variants do pairwise inner (or outer) product between all fields and concatenate the outputs as first layer of DNN. 
In Wide \& Deep~\cite{cheng2016wide}, DNN is joined with a linear model, where the inputs of the linear model are handcrafted high-order features by experts, as shown in Figure \ref{fig:related}(b). To reduce the human intervention, we use original features in the linear part instead of the handcrafted features.
Most of the existing deep tabular data classification models follow the structure of Wide \& Deep.
Neural Factorization Machines~\cite{he2017neural} developed bi-interaction pooling of the fields as the first layer of DNN.
Deep factorization machine~\cite{guo2017deepfm} uses FM~\cite{rendle2010factorization} to replace the linear part of Wide \& Deep to reduce the dependence on feature engineering, while AutoInt~\cite{song2018autoint} replaces the linear part with a multi-head self-attentive neural network. 
As we can see from the above methods, the information that features within each field belong to the same field and the non-linear interactions between operations are overlooked.

Recently, how to leverage behavior sequences of users has also drawn much attention~\cite{bst_kdd2019,feng2019deep}.
\cite{bst_kdd2019} proposes to use behavior sequence transformer, while \cite{hpmn_sigir2019} uses hierarchical periodic memory network.
Deep Interest Evolution Network~\cite{dien_aaai2019} and Deep Session Interest Network~\cite{feng2019deep} also take session information into consideration.
Since not all of our customers' data have behavior sequence information, we will focus on the scenario without this information, and methods in this area are not further discussed.

\section{Network On Network}\label{sec:proposedmethod}
\begin{figure}[!ht]
    \centering
    \includegraphics[width=0.5\textwidth]{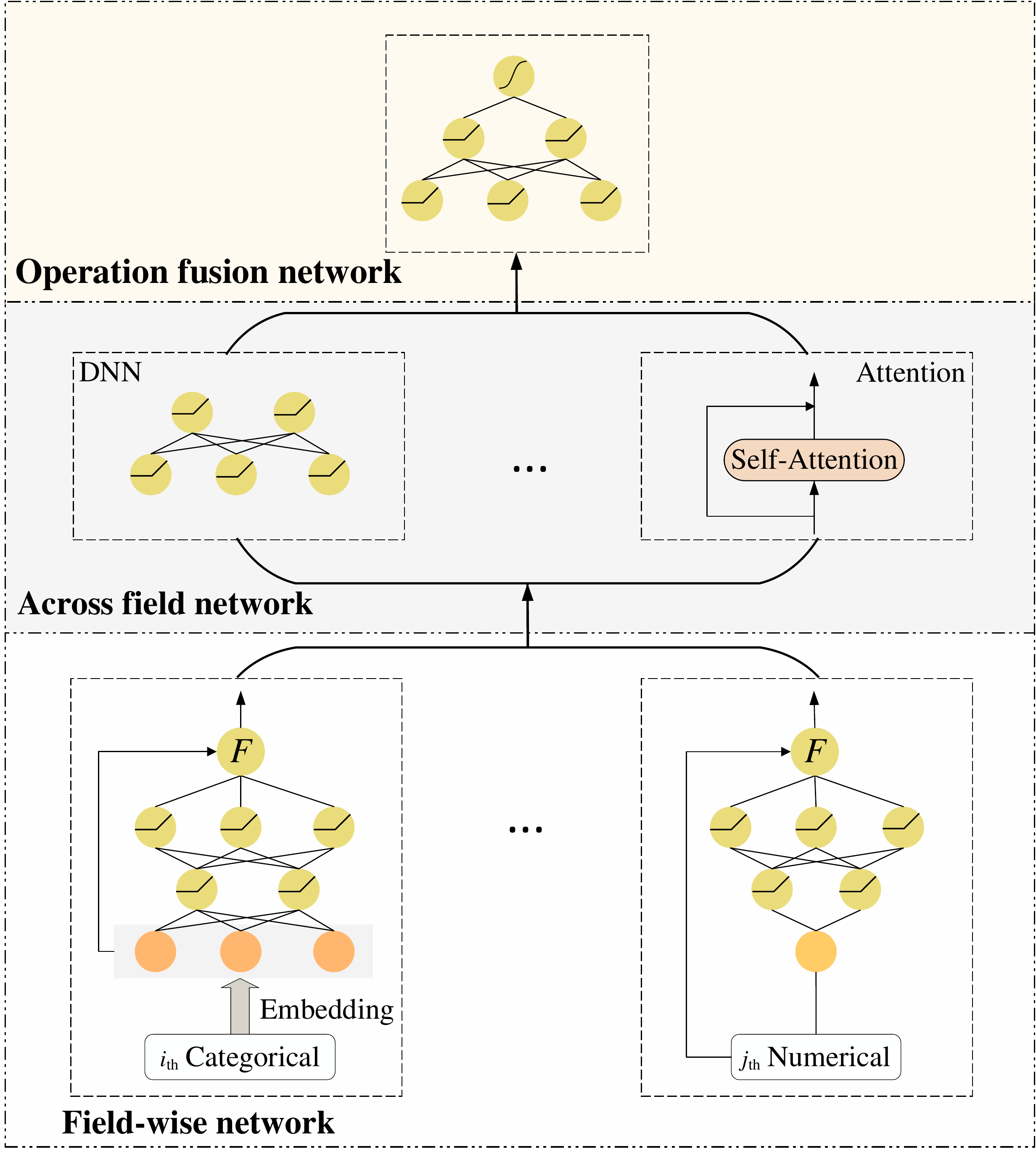}
    \caption{Overall structure of Network On Network (NON).}
    \label{fig:framework} 
\end{figure}

To better capture the intra-field information that features inside each field belong to the same field and the no-linear interactions between various operations, we propose \textit{Network On Network} (NON). Moreover, the operations used in NON are data-driven, which means different datasets may use different operations.
The overall structure of NON is shown in Figure \ref{fig:framework}. 
NON consists of three parts, including field-wise network at the bottom, across field network in the middle and operation fusion network on the top.
In the field-wise network, features belong to the same field share a neural network (NN) to capture the intra-field information for each field, i.e. each field corresponding to a unique NN.
In the across field network, a variety of operations are employed to model potential interactions between fields, such as LR to model the linear interactions explicitly and multi-layer neural network to model high-order interactions implicitly.
In the operation fusion network, the high-level feature representations learned by different operations in the across field network are fused by a DNN to get the final prediction. 

Due to deep stacked structure of NON, it may lead to even worse results.
Inspired by GoogLeNet~\cite{szegedy2015going}, we introduce DNN with auxiliary losses to ease this problem, where 
auxiliary classifiers are added to every layer of DNN.
Note that GoogLeNet (22 layers deep) only add two auxiliary classifiers, while we add it to every layer of DNN.

In the following, we first represent the structure of NON, and then its training technique, i.e. DNN with auxiliary losses is described, finally time complexity of NON is analyzed.

\subsection{The structure of NON}
\subsubsection{Field-wise network}
While most of the existing deep tabular data classification models do not take full advantage of the intra-field information, we propose to use field-wise network to capture this information.
To be specific, each field owns a DNN, and for features in each field, their embeddings are fed into their DNN first. The parameters in the DNNs are used to store the intra-field information.
Let $\bm{e_i}$ and $ \textup{DNN}_{i} $ be the embedding and DNN of the $i$-th field respectively, then the output is
\begin{equation}\label{eq:field_dn}
	\bm{e^{'}_{i}} = \textup{DNN}_{i} \left( \bm{e_i} \right) .
\end{equation}
In practice, if multiple fields have the same structure of the DNN in Equation \ref{eq:field_dn}, they can be computed in parallel to speedup.
We can stack inputs and weights of each layer within their DNNs, and then use matrix multiplication to calculate the outputs once and for all.
Suppose we have $c$ fields (corresponding to $c$ DNNs which have the same structure), we will take a specific layer for all the $c$ DNNs as an example.
Denote $\bm{X_i}  \in \mathbb{R}^{b \times d_1}$ and $\bm{W_{i}} \in \mathbb{R}^{d_1 \times d_2}$ as the batch of inputs and weights of the $i$-th DNN for the given layer, where $b$ is the mini-batch size,  $d_1$ and $d_2$ are inputs and outputs dimension.
We can stack the inputs and weights like
\begin{equation*}
	\begin{aligned}
	 	\mathbf{X} &= \textup{stack} \left( \left[ \bm{X_1}, \bm{X_2}, \cdots, \bm{X_c} \right] \right) \in \mathbb{R}^{c \times b \times d_1} \\
	 	\mathbf{W} &= \textup{stack} \left( \left[ \bm{W_1}, \bm{W_2}, \cdots, \bm{W_c} \right] \right) \in \mathbb{R}^{c \times d_1 \times d_2}.
	 \end{aligned}
\end{equation*}
Finally we can compute the output by
\begin{equation*}
	\mathbf{X}^{'} = \textup{ReLU}\left( \textup{matmul} \left( \mathbf{X}, \mathbf{W} \right) + \mathbf{b} \right),
\end{equation*}
where $\mathbf{X}^{'} \in \mathbb{R}^{c \times b \times d_2}$ is the output, $\mathbf{b} \in \mathbb{R}^{c \times 1 \times d_2}$ is bias term, and `matmul' is batch matrix multiplication both supported by Tensorflow\footnote{https://www.tensorflow.org/api\_docs/python/tf/linalg/matmul} \cite{tensorflow2015} and PyTroch\footnote{https://pytorch.org/docs/1.1.0/torch.html?\#torch.matmul} \cite{pytroch2019}.

While the outputs of field-wise network can be fed into upper networks directly, we can also refine the outputs by the original embeddings using
\begin{equation*}
	\bm{\hat{e}_{i}} = F\left( \bm{e^{'}_{i}}, \bm{e_{i}} \right),
\end{equation*}
where $\bm{e^{'}_{i}}$ and $\bm{e_{i}}$ are the same as Equation \ref{eq:field_dn}, $F$ is a function, which can be concatenation, element-wise product, or more complex function like gating mechanism~\cite{gers1999learning,chung2014empirical}.

\subsubsection{Across field network}
Interactions between fields are central to classification performance, and various operations~\cite{rendle2010factorization,cheng2016wide,guo2017deepfm,he2017neural,lian2018xdeepfm,song2018autoint} have been proposed to model the interactions in different manners.
In the across field network, multiple operations are adopted to learn distinct types of interactions between different fields. Typical operations are: 
\begin{itemize}
	\item Linear model \quad Though linear models like logistic regression (LR) are simple, they are quite stable as easy to train. Moreover, user can add human-designed features to linear part to further enhance the performance, thus they are adopted in various models \cite{cheng2016wide,lian2018xdeepfm}. In this paper, LR is used and the original features are fed to LR. 
    
    \item vanilla DNN \quad DNN can learn high order interactions among different fields, and almost all the existing methods \cite{cheng2016wide,guo2017deepfm,lian2018xdeepfm} include this module. DNN is also the requisite operation in NON.
    
    \item self-attention \quad  Attention mechanism~\cite{luong2015effective,vaswani2017attention} have achieved great success in natural language processing and computer vision, and it is also introduced to solve tabular classification problem \cite{song2018autoint}. 
    When applying on tabular data, attention mechanism will give relevant input higher weight to form more meaningful representation. 
    In this paper, multi-head self-attentive network \cite{vaswani2017attention} is chosen as one of our candidate operations.
    
    \item FM and Bi-Interaction \quad Factorization Machines (FMs)~\cite{rendle2010factorization} is an effective way to model second-order field interactions explicitly. Bi-Interaction~\cite{he2017neural} generalize the formalization of FM. It is a pooling operation that converts a set of embedding vectors to one vector, where element-wise product is adopted for each pair of embedding vectors. The calculation of Bi-Interaction is
    \begin{equation*}
    	\bm{v} = \sum_{i}^{m} \sum_{j}^{m} x_i  \bm{e_i}  \odot x_j \bm{e_j} ,
    \end{equation*}
    where $\odot$ denotes element-wise product of two vectors, $x_i$ is the feature value, $\bm{e_i}$ is the embedding of $i$-th field, $\bm{v}$ is the output vector with the same dimension with $\bm{e_i}$.
    We choose Bi-Interaction as the candidate operations instead of FM because Bi-Interaction can recover the FM model exactly\cite{he2017neural}.
\end{itemize}

The outputs of operations like self-attention, Bi-Interaction and DNN are vectors, and only the output of LR is scalar.
As we can see, the architecture of NON makes it compatible with most of the existing deep classification models.
Furthermore, unlike previous works use a predefined combination of operations, the operations in NON except DNN are treated as hyper-parameters and will be determined by the input data, i.e. the operations used in NON are data-driven.

\subsubsection{Operation fusion network}
For existing methods shown in Table \ref{tab:proposed_ops}, the outputs of different operations are first concatenated, and then weighted sum to get the final prediction before Sigmoid function. Because of intrinsic linearity of weighted sum, the non-linear interactions between the operations are ignored.
To better capture the interactions between operations, NON employs a DNN to further fuse outputs of different operations.
To be specific, the input of operations fusion network is
\begin{equation*}
	\bm{x_{\textup{ofn}}} = \textup{concat} \left( \left[ \bm{o}_{1}, \bm{o}_{2}, \cdots, \bm{o}_{k}\right] \right)  \in \mathbb{R}^{\sum_{i}d_{i}},
\end{equation*}
where $\bm{o}_{i} \in \mathbb{R}^{d_{i}}$ denotes the output of operation $i$,  whose dimension is $d_{i}$. 
Then, we can get the final prediction as
\begin{equation*}
y^{'}=\textup{DNN}_{\textup{ofn}} \left(\bm{x_{\textup{ofn}}}\right),
\end{equation*}
where $\textup{DNN}_{\textup{ofn}}$ is the DNN employed in the operation fusion network.
Operation fusion network generalizes existing methods, and they can be seen as a special case of operation fusion network when $\textup{DNN}_{\textup{ofn}}$ has only one layer.
Benefits from the expressiveness of DNN, the interactions between different operations are fully exploited.
To avoid the computation burden caused by this component, we recommend the depth and units of hidden layer are no larger than 2 and 64, respectively.

\subsection{DNN with auxiliary losses}
\begin{figure}[!ht]
    \centering
    \includegraphics[width=0.36\textwidth]{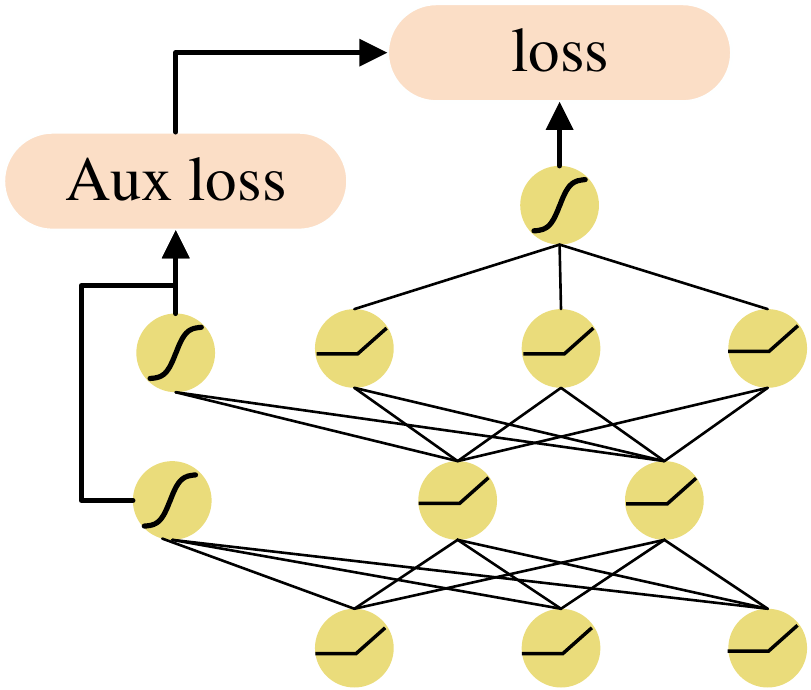}
    \caption{The structure of DNN with auxiliary losses.}\label{fig:auxiliarydnn}
\end{figure}

Because of the deep stacked structure of NON, it may lead to even worse results. Inspired by GoogLeNet~\cite{szegedy2015going}, we introduce DNN with auxiliary losses, as shown in Figure \ref{fig:auxiliarydnn}, to make NON easier to train. Intuitively, auxiliary losses make the intermediate layers more discriminative. In DNN with auxiliary losses, each layer corresponds to an auxiliary loss. Suppose $\bm{h}_{i}$ is the hidden state of $i$-th layer in DNN, then the auxiliary loss of layer $i$ is defined as
\begin{equation}\label{eq:aux}
    \ell_{aux}^{i} = \ell \left( \textup{sigmoid}\left( {{\mathbf{W}}_{aux}}^T_{i}  \bm{h}_{i} \right), y \right),
\end{equation}
where $y$ is the label of an instance, $\ell(y', y)$ is cross entropy between $y'$ and $y$, ${{\mathbf{W}}_{aux}}^T_{i} $ are trainable variables. Then, the loss function of NON is
\begin{equation*}\label{eq:loss}
    \ell = \ell(y', y) +
    \alpha \sum_{i}{ \ell_{aux}^{i} } + \gamma \left \| \mathbf{W} \right \|, 
\end{equation*}
where $y'$ is final prediction of the top part, $\alpha$ and $\gamma$ are hyper-parameters, and $\left \| \mathbf{W} \right \|$ is the $\textup{L}_2$-norm of the network weights. In practice,  we can use different $\alpha$ for auxiliary loss of different layer and decay $\alpha$ when training.  
We only use the auxiliary loss in the training process, i.e. the network degrades to vanilla DNN when doing inference.

\subsection{Time complexity analysis}
Suppose the dimension of feature embedding is $d$, the time complexity of the field-wise network is $O(md^2L_f)$, where $m$ is the number of fields, $L_f$ is number of layers. For the across field network, its time complexity is determined by the time complexity of operations: the time complexity of LR is $O(m)$; the time complexity of DNN is $O\left(H_a^2L_a\right)$, where $H_a$ and $L_a$ is the average layer size and number of layers; the time complexity of Bi-Interaction and FM is $O( m^2d )$; the time complexity of the attention based operation is $O\left( m^2N_{h}d'+mN_{h}d'd \right)$, where $N_{h}$ is number of head, $d'$ is the attention embedding dimension.
The time complexity of the operation fusion network is $O\left( H_o^2L_o \right)$, where $H_o$ and $L_o$ is the average layer size and number of layer, respectively. The time complexity of NON is dominated by the across field network,
and one can reduce the time complexity of NON by choosing light-weight operations in the across field network.

\section{Numerical experiment}\label{sec:exp}
As the key contributions of this work are the field-wise network and the design paradigm of NON, we conduct extensive numerical experiments to answer the following questions:
\begin{itemize}
    \item[\textbf{Q1}] How does the design paradigm of NON perform?
    \item[\textbf{Q2}] How does NON perform as compared to state-of-the-art methods?
    \item[\textbf{Q3}] What are the most suitable operations for different datasets?
    \item[\textbf{Q4}] Can the field-wise network in NON capture the intra-field information effectively? 
\end{itemize}
In the following, we first describe the setup of the experiments and answer these questions.

\subsection{Experiment setup}
\subsubsection{Datasets} 
The numerical experiments are conducted on six real-world datasets, and three of them are provided by our customers. Statistics of the datasets are summarized in Table \ref{tab:expdata}.

\textbf{Criteo\footnote{https://www.kaggle.com/c/criteo-display-ad-challenge}:} This is a  week's display advertising data shared by CriteoLab for ad click-through rate estimation, and it is also widely used in many research papers. The data consists of 45 million users' click records, which contains 13 numerical fields and 26 categorical fields. 

\textbf{Avazu\footnote{https://www.kaggle.com/c/avazu-ctr-prediction}:} This is a data provided by Avazu to predict whether a mobile ad will be clicked. It contains 40 million records of users' mobile behavior with 23 categorical fields.

\textbf{Movielens\footnote{https://grouplens.org/datasets/movielens/}:} This data is rating data from the MovieLens web site collected by GroupLens Research. It is a data about 20 million records of users' ratings on movies. The data has 1 numerical field and 3 categorical fields.

\textbf{Talkshow:} This data is provided by our customer, which is a talk show company. The task is to predict whether a user will watch the shows and how long he/she will watch. The data has more than 2 million records, which contains 9 numerical fields and 21 categorical fields.

\textbf{Social:} This data comes from a customer which provides social network service. The task is to recommend new friends to its users to increase daily active users. The data consists of about 2 million records which contains 57 numerical fields and 18 categorical fields.

\textbf{Sports:} This data is provided by our customer who runs a fitness app. When users do exercises with this app, the app needs to push some videos and musics that users might be interested. The data has about 3 million records, which contains 34 numerical fields and 28 categorical fields.

\begin{table}[ht]
	\caption{The statistics of the datasets (\#Num.: number of numerical fields; \#Cate.: number of categorical fields; \#Val.: number of feature values in all the categorical fields).}
	\label{tab:expdata}
    \centering
    \renewcommand\tabcolsep{4pt}
    \begin{tabular}{lrrrrr}
      \toprule
      \multirow{2}{*}{Name} & \multicolumn{2}{c}{\#Samples} &  \multicolumn{3}{c}{\#Fields} \\
        \cmidrule(r){2-3} \cmidrule(r){4-6}
       & Training & Testing & \#Num. & \#Cate. & \#Val.  \\
      \midrule
      Criteo & 41,256K & 4,548K & 13 & 26 & 33,762K  \\
      Avazu & 36,386K & 4,042K & 0 & 23 & 1,544K  \\
      Movielens & 16,000K & 4,000K & 1 & 5 & 155K  \\
      Talkshow & 1,888K & 1,119K & 9 & 21 & 366K  \\
      Social & 1,153K & 796K & 57 & 18 & 1,895K  \\
      Sports & 2,641K & 719K & 34 & 28 & 4,181K \\ 
      \bottomrule
    \end{tabular}
\end{table}

\subsubsection{Data preparation}
The missing values and features with frequency less than threshold $T$ in the categorical fields are treated as `unkown'. In this paper, a typical value of 5 is selected for $T$.
Numerical features are normalized, and the missing values in numerical fields are assigned with 0.
As shown in Table \ref{tab:expdata}, the datasets are split into training and test, while 20\% of training data is used for validation.

\subsubsection{Evaluation metrics}
We use AUC (Area Under the ROC curve) as the metric. As reported in  \cite{cheng2016wide,guo2017deepfm}, 0.275\% improvement in off-line AUC can lead to 3.9\% in online CTR, which in turn brought extra millions of dollars revenue. i.e small improvement of off-line AUC can lead to significant increase in online business revenue and hence great commercial benefits.

\subsubsection{Baselines}
We compare the proposed method with:
\begin{itemize}
  \item FFM~\cite{juan2016field}: it is an official implementation\footnote{https://github.com/ycjuan/libffm} of FFM released by Yuchin Juan, which uses field-aware factorization techniques to model second order feature interactions.
  \item DNN~\cite{lecun2015deep}: it is a vanilla deep neural network, which can learn high order interactions among different fields implicitly. It is used as the baseline.
  \item Wide \& Deep~\cite{cheng2016wide}: it is DNN joined with a linear model. While the linear part is fed with handcrafted high-order features by experts in \cite{cheng2016wide}, we use the original features instead for fair comparison.
  \item NFM~\cite{he2017neural}: it stacks a DNN on top of FM, where DNN is used to implicitly capture the higher feature interactions. For fair comparison, NFM without pre-training is adopted in this paper.
  \item xDeepFM~\cite{lian2018xdeepfm}: it combines DNN and CIN to build one unified model, where CIN is used to learn feature interactions explicitly. Both DNN and CIN are fed with all the feature fields.
  \item AutoInt~\cite{song2018autoint}: it consists of a multi-head self-attentive neural network with residual connections and DNN, where the self-attention neural network with residual connections uses self-attentive mechanism to learn feature interactions explicitly. Both the two parts are fed with all the feature fields.
\end{itemize}
Among them, NFM, xDeepFM and AutoInt are the state-of-the-art methods.

\subsubsection{Reproducibility}\label{sec:reproducibility}
The FFM used is provided by its authors, and we implement other method using Tensorflow\footnote{https://github.com/tensorflow/tensorflow}. For fair comparison, the hyper-parameters of all models except FFM in the experiment use the same search space. The optimizer is Adagrad with mini-batch size of 256; the learning rate  $\in[0.05, 0.5]$; the embedding dimension $d 
\in \{8, 16, 32\cdots, 128\}$; the layer size of DNN $\in \{2048, 1024, 512 \cdots 64\}$, while the number of layers of DNN $\le 4$; the layer size of the field-wise network $\in \{3.0,2.0,1.5,1.0,0.5\}\times d$, while the number of layers $\le 4$; the auxiliary cofficient $\alpha \in [0.1, 1.0]$; the $\textup{L}_2$-norm cofficient $\gamma \in [1e-5, 1e-4]$; the space of other hyper-parameters are consistent with the setting in baselines.
Optional operations are LR, FM, Bi-Interaction and multi-head self-attention, and DNN is always required.
For FFM, the dimension of embedding is 4, and learning rate $\in [0.01, 0.5]$.
We use random search~\cite{bergstra2012random} and run 60 times for each method.

\begin{figure}[!ht]
  \centering 
  \includegraphics[width=0.48\textwidth]{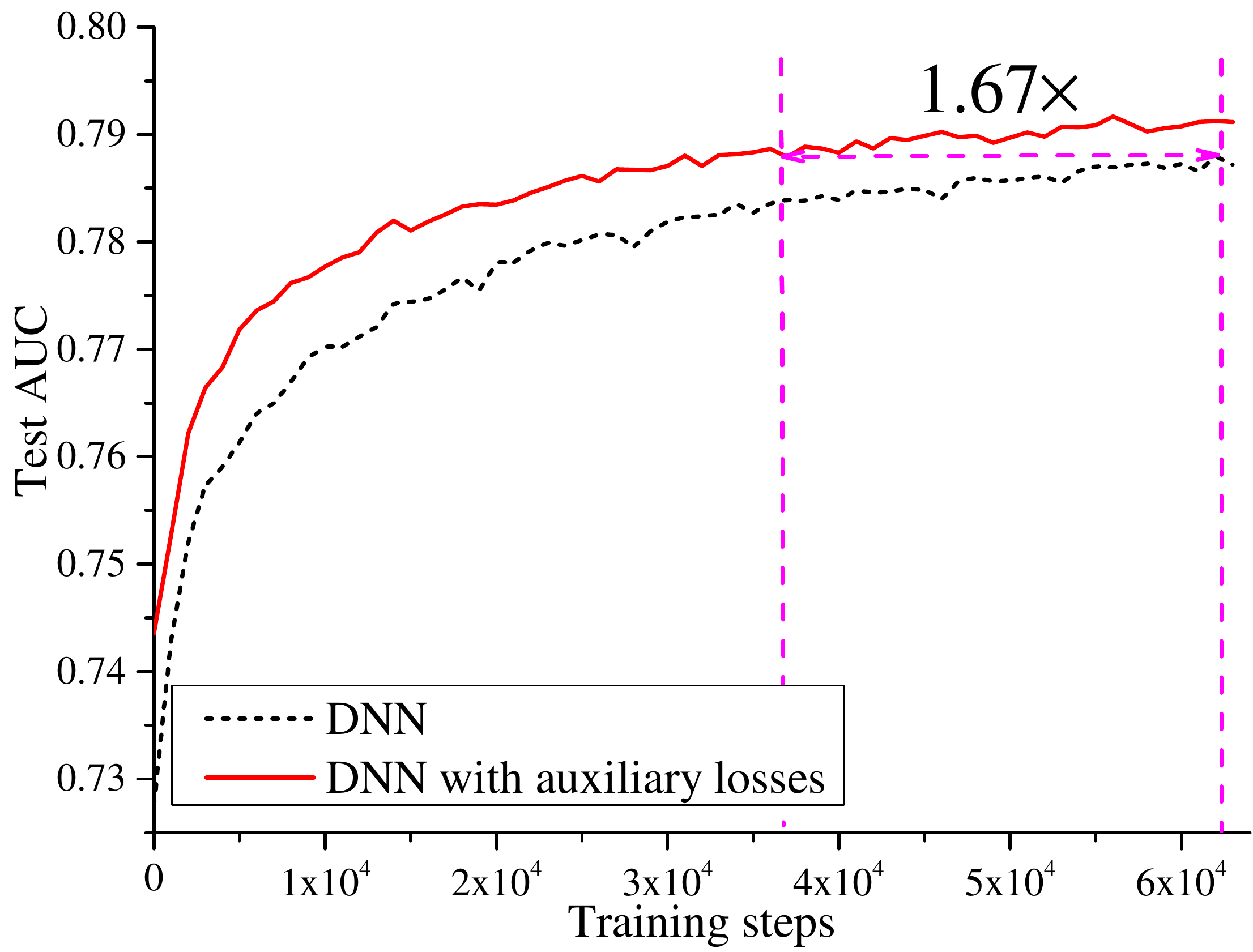}
  \caption{The test AUC on a subset of Criteo in the training process using DNN and DNN with auxiliary losses.}\label{fig:subcriteo}
\end{figure}

\begin{table*}[!htb] 
	\caption{Ablation study of NON. (DNN with aux.: DNN with auxiliary losses; field.: field-wise network with a upper DNN; across.: across field network; operation.: operation fusion network. imp.: test AUC improvement compared with DNN).}
	\centering
	\renewcommand\tabcolsep{7pt}
	\begin{tabular}{lcccccccc}
		\toprule 
		\multirow{2}{*}{Dataset} &  \multirow{2}{*}{DNN} &  \multirow{2}{*}{DNN with aux.}   & \multicolumn{2}{c}{field.} & \multicolumn{2}{c}{field. + across.} &  \multicolumn{2}{c}{NON(field. + across. + operation.)}\\ 
		\cmidrule(r){4-5} \cmidrule(r){6-7} \cmidrule(r){8-9} 
		& & & AUC & imp. (\%) & AUC & imp. (\%)  & AUC & imp. (\%)  \\
		\midrule 
		Criteo         & 0.8063    & 0.8084     & 0.8094  & 0.38 & 0.8108 & 0.56 & \textbf{0.8115} & \textbf{0.64} \\
		Avazu         & 0.7763    & 0.7809     & 0.7821  & 0.75 & 0.7827 & 0.82 & \textbf{0.7838} & \textbf{0.97} \\
		Movielens   & 0.6988    & 0.7018     & 0.7035 & 0.67 & 0.7036 & 0.69 & \textbf{0.7057} & \textbf{0.99} \\
		Talkshow    & 0.8451    & 0.8519     & 0.8525  & 0.88 & 0.8521 & 0.83 & \textbf{0.8533} & \textbf{0.97} \\
    Social         & 0.6969    & 0.6998     & 0.6996  & 0.39 & 0.7018 & 0.70 & \textbf{0.7032} & \textbf{0.90} \\
    Sports         & 0.8506    & 0.8526     & 0.8534  & 0.33 & 0.8533 & 0.32 & \textbf{0.8561} & \textbf{0.65} \\
		\bottomrule 
	\end{tabular}
	\label{tab:auxdnn}
\end{table*}

\begin{table*}[!htb]  

	\caption{Comparison with state-of-art methods (imp.: the test AUC improvement compared with DNN).}
	\label{tab:hfn}
	
    \centering
    \renewcommand\tabcolsep{4pt}
    \begin{tabular}{lcccccccccccc}
      \toprule 
      \multirow{2}{*}{Dataset} &\multirow{2}{*}{DNN} &  \multirow{2}{*}{FFM} &  \multicolumn{2}{c}{Wide \& Deep} & \multicolumn{2}{c}{NFM}  & \multicolumn{2}{c}{xDeepFM} & \multicolumn{2}{c}{AutoInt}  & \multicolumn{2}{c}{NON} \\
      \cmidrule(r){4-5} \cmidrule(r){6-7} \cmidrule(r){8-9} \cmidrule(r){10-11} \cmidrule(r){12-13}
      &   &   & AUC & imp. (\%) &  AUC & imp. (\%) & AUC & imp.(\%) & AUC & imp.(\%) & AUC & imp.(\%)  \\
      \midrule
      Criteo     & 0.8063 & 0.8016 & 0.8052 & -0.14& 0.8025 & -0.47 & 0.8102 & 0.48 & 0.8051 & -0.15  &   \textbf{0.8115} & \textbf{0.64}\\
      Avazu      & 0.7763 & 0.7830 & 0.7773 & 0.13 & 0.7787 &  0.31 & 0.7792 &  0.37 & 0.7775  & 0.16 & \textbf{0.7838} & \textbf{0.97}\\
      Movielens  & 0.6988 & 0.6930 & 0.6991 & 0.04 & 0.6991 &  0.04  & 0.6994 & 0.09 & 0.7004  & 0.22 &  \textbf{0.7057} & \textbf{0.99} \\
      Talkshow   & 0.8451 & 0.8253 & 0.8491 & 0.49 & 0.8200 &  -2.97 & 0.8502 & 0.60 & 0.8470 & 0.22 &    \textbf{0.8533} & \textbf{0.97} \\
      Social     & 0.6969 & 0.6654 & 0.6986 & 0.24 & 0.6952 &  -0.24 & 0.7015 & 0.66 & 0.7015 & 0.66  &     \textbf{0.7032} & \textbf{0.90}\\
      Sports    & 0.8506 & 0.8100  & 0.8512 & 0.07 & 0.8434 &  -0.08 & 0.8501 & -0.06 & 0.8512 & 0.07  &  \textbf{0.8561} & \textbf{0.65} \\
      \bottomrule 
    \end{tabular}
	\vspace{-3pt}
\end{table*}

\subsection{Study of design paradigm of NON (Q1)}
\label{subsec:design_paradigm}
\textbf{Design paradigm of NON improves generalization.} \quad
Due to the deep stacked structure of NON, it may lead to even worse results. To alleviate this problem, we introduce the training technique that adding auxiliary losses to every layer of DNN, as shown in Figure \ref{fig:auxiliarydnn}. 
Benefits from the short path of auxiliary losses, vanishing gradient can be effectively alleviated and therefore training efficiency is improved~\cite{szegedy2015going}. As shown in Figure \ref{fig:subcriteo}, in the training process on a subset of Criteo dataset, DNN with auxiliary losses can speed up the training process about 1.67$\times$ compared with DNN when achieving the same test AUC.
As shown in Table \ref{tab:auxdnn}, DNN with auxiliary losses can also improve the performance of the model.
The results show the effectiveness of the training technique and all the subsequent results in Table \ref{tab:auxdnn} are obtained with this technique.

We further demonstrate the effectiveness of the design paradigm of NON through ablation studies,
and the results are shown in Table \ref{tab:auxdnn}.
From the results, we can see that:
1) NON outperforms all the other methods on all the datasets. This observation demonstrates the generality and superiority of our approach.
2) When integrating DNN with field-wise network, the performance grows at most of the time. The improvements indicate the intra-field information captured by the field-wise network does help to improve model performance.
3) When the components of NON are stacked up, the performance grows consistently. In the across field network, various operations are explored and the gains in the results show its strength. By capturing non-linear interactions between distinct operations, the performance are further improved, which indicates the need of operation fusion network. These results further demonstrate the superiority of our design paradigm of NON. 
\begin{figure*}[!ht]
  \makeatletter
  \renewcommand{\@thesubfigure}{\hskip\subfiglabelskip}
  \makeatother
      \centering
      \subfigure[Criteo]{
        \label{fig:opcriteo} 
        \includegraphics[width=0.32\textwidth]{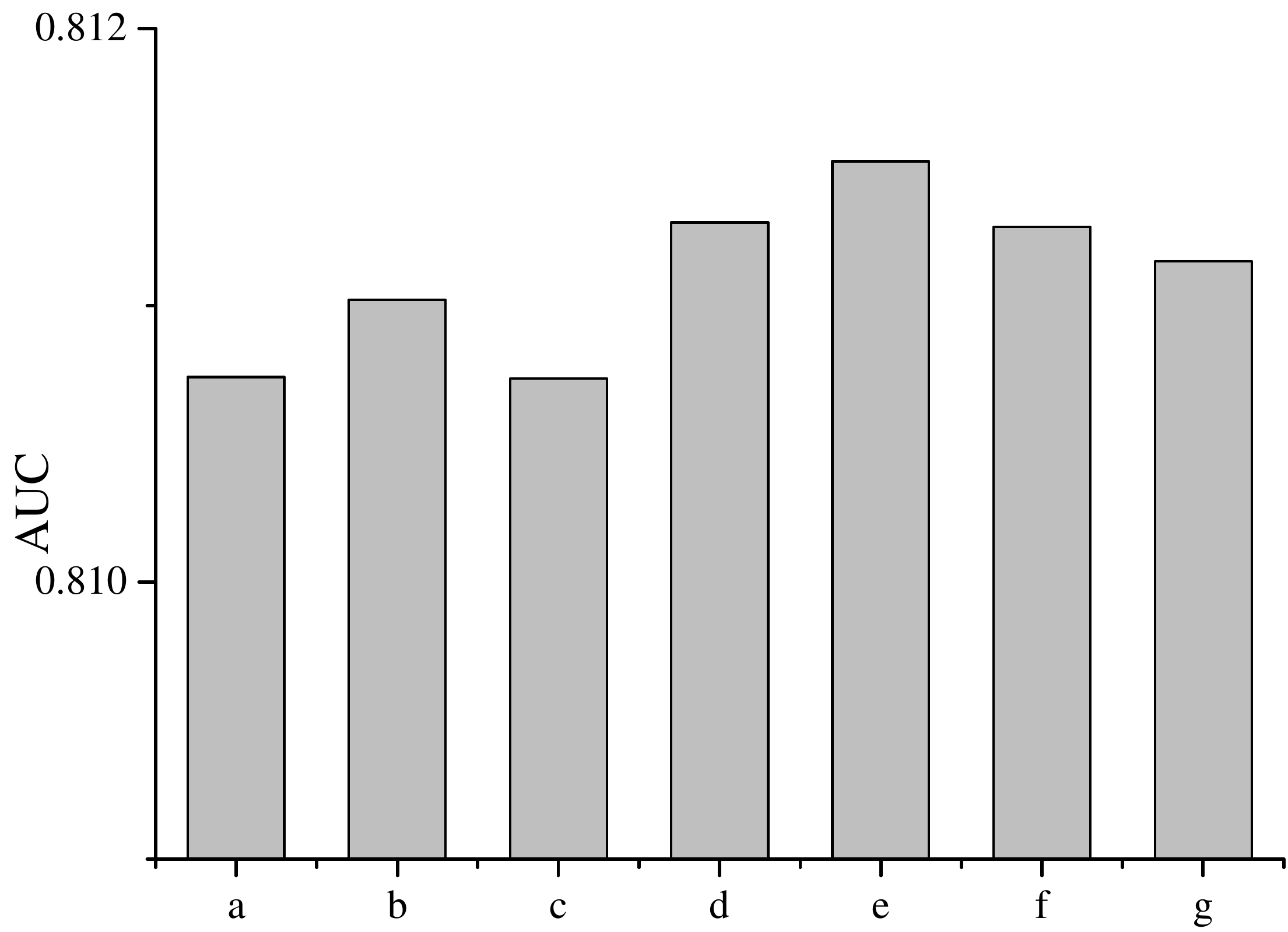}
      }
      \subfigure[Avazu]{
        \label{fig:opavazu} 
        \includegraphics[width=0.32\textwidth]{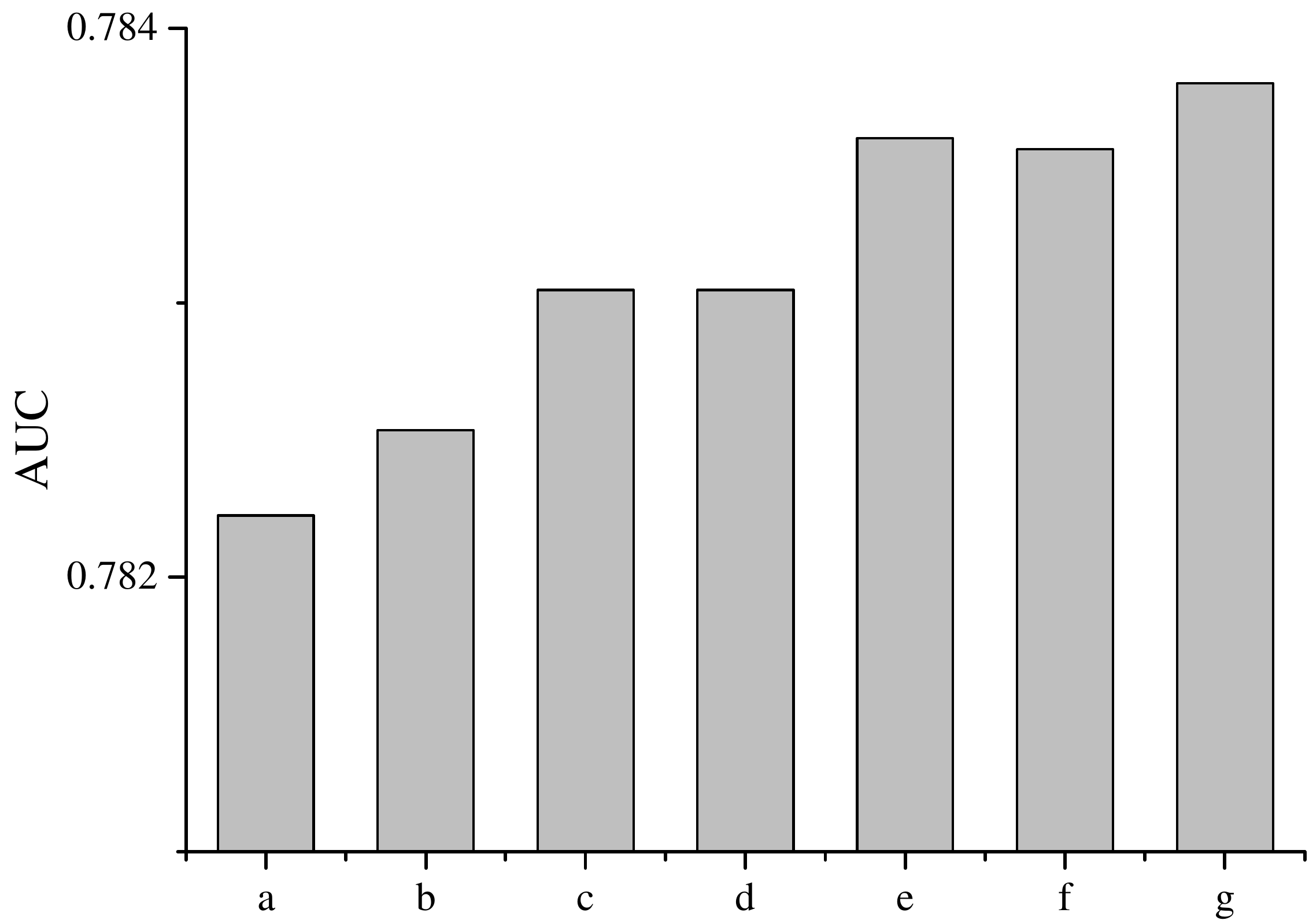}
      }
      \subfigure[Movielens]{
        \label{fig:opml} 
        \includegraphics[width=0.32\textwidth]{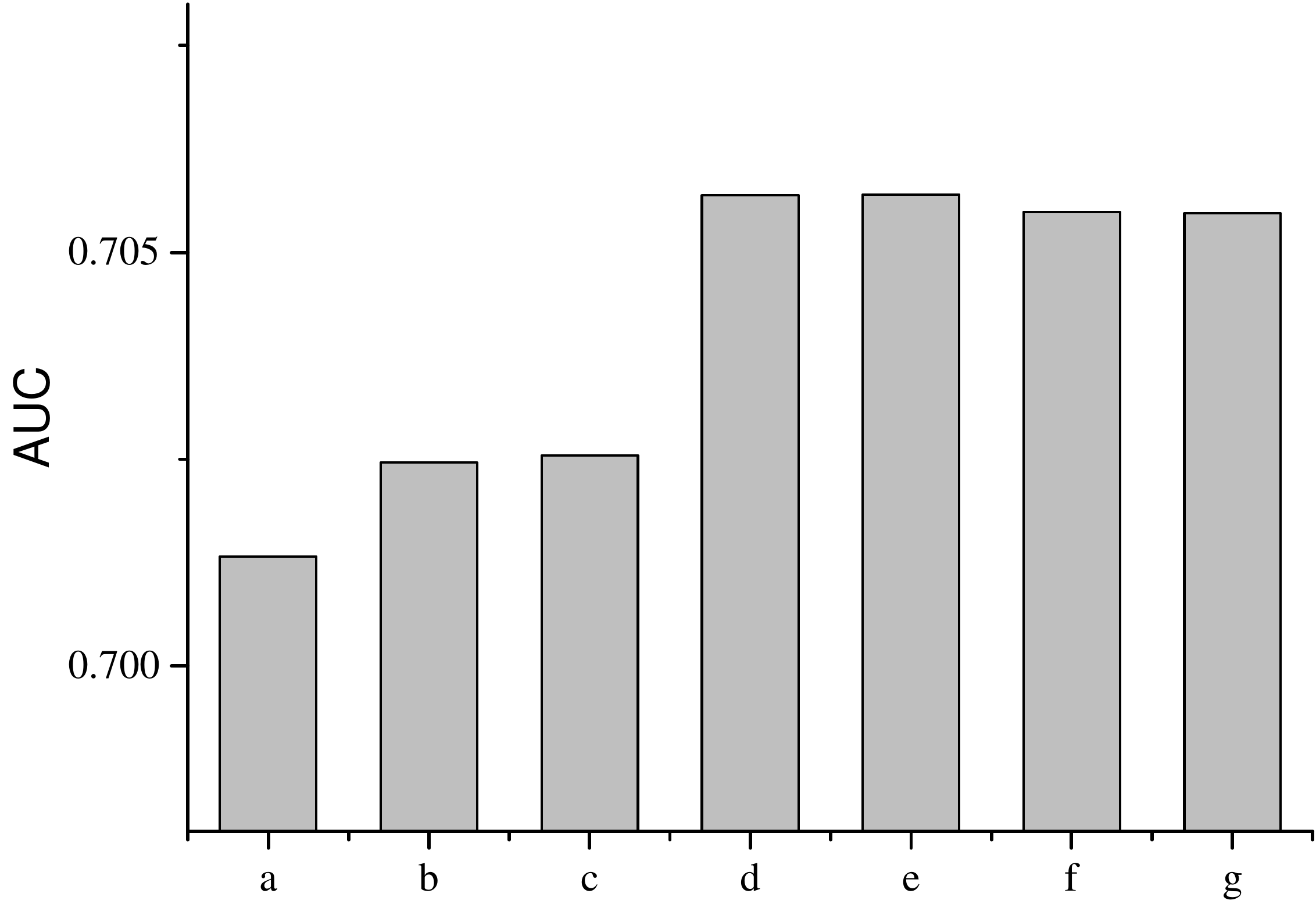}
        }
      \subfigure[Talkshow]{
      \label{fig:opluoji} 
      \includegraphics[width=0.32\textwidth]{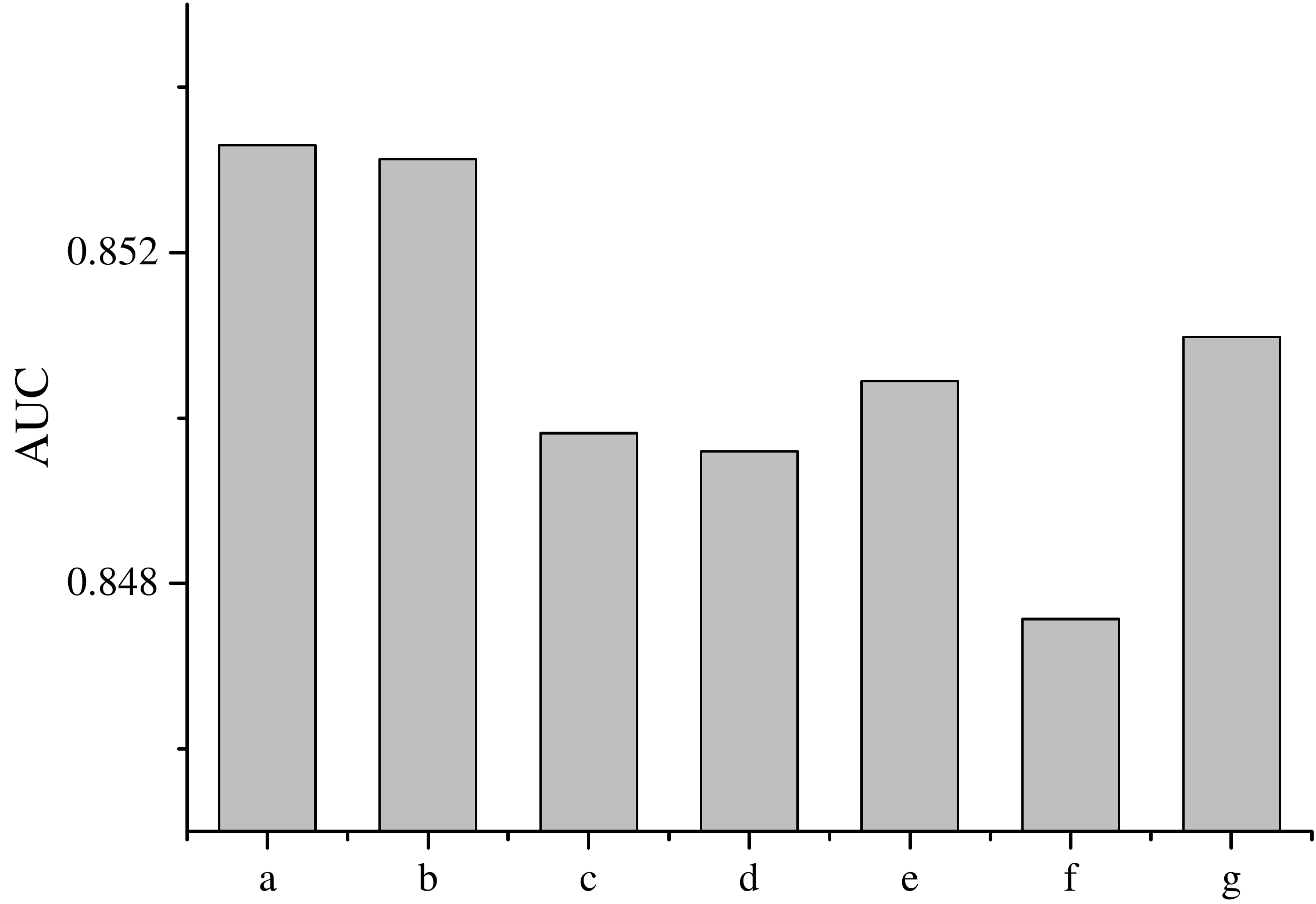}
      }
      \subfigure[Social]{
        \label{fig:opmyhug} 
        \includegraphics[width=0.32\textwidth]{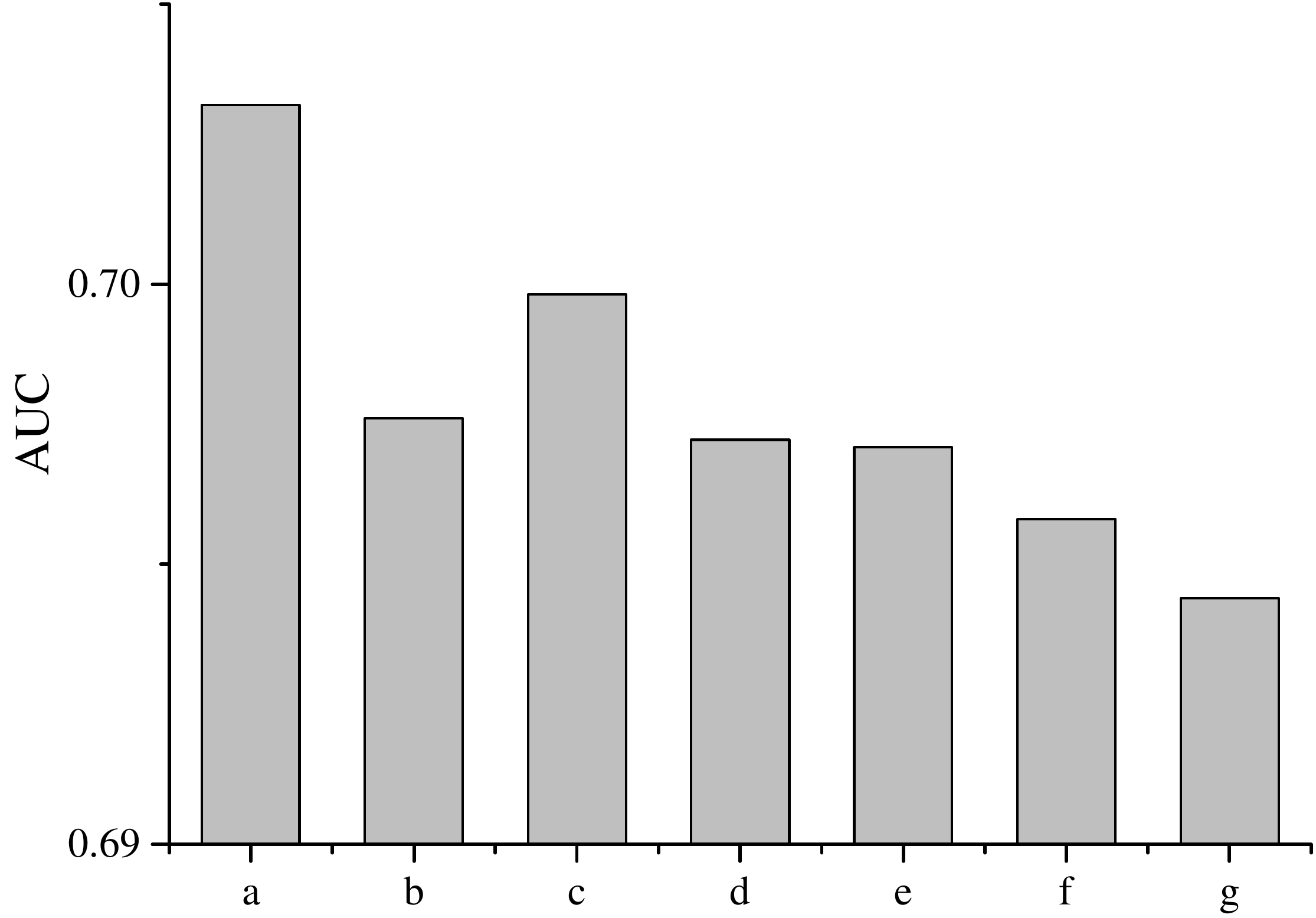}
      }
      \subfigure[Sports]{
        \label{fig:opsports} 
        \includegraphics[width=0.32\textwidth]{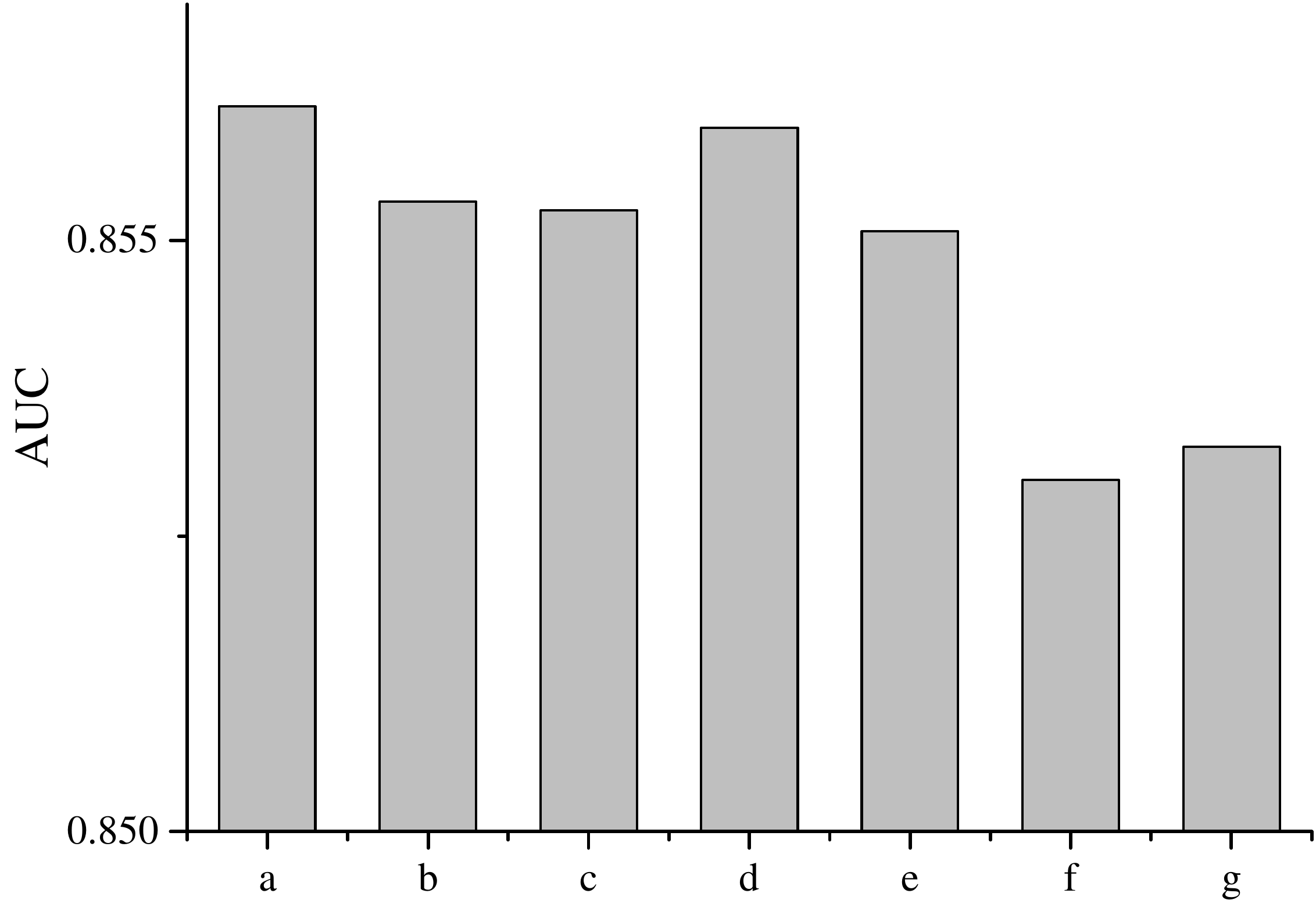}
      }

    \caption{AUC of NON with different combinations of operation on six datasets. (a: DNN \& LR; b: DNN \& Bi-Interaction; c: DNN \& Attention; d: DNN \& LR \& Bi-Interaction; e: DNN \& LR \& Attention; f: DNN \& Bi-Interaction \& Attention; g: DNN \& LR \& Bi-Interaction \& Attention.)}
    \label{fig:opcombines}
  \end{figure*}
  
  \begin{figure*}[!ht]
    \makeatletter
    \renewcommand{\@thesubfigure}{\hskip\subfiglabelskip}
    \makeatother
        \centering
        \subfigure[Criteo]{
          \label{fig:embcriteo} 
          \includegraphics[width=0.3\textwidth]{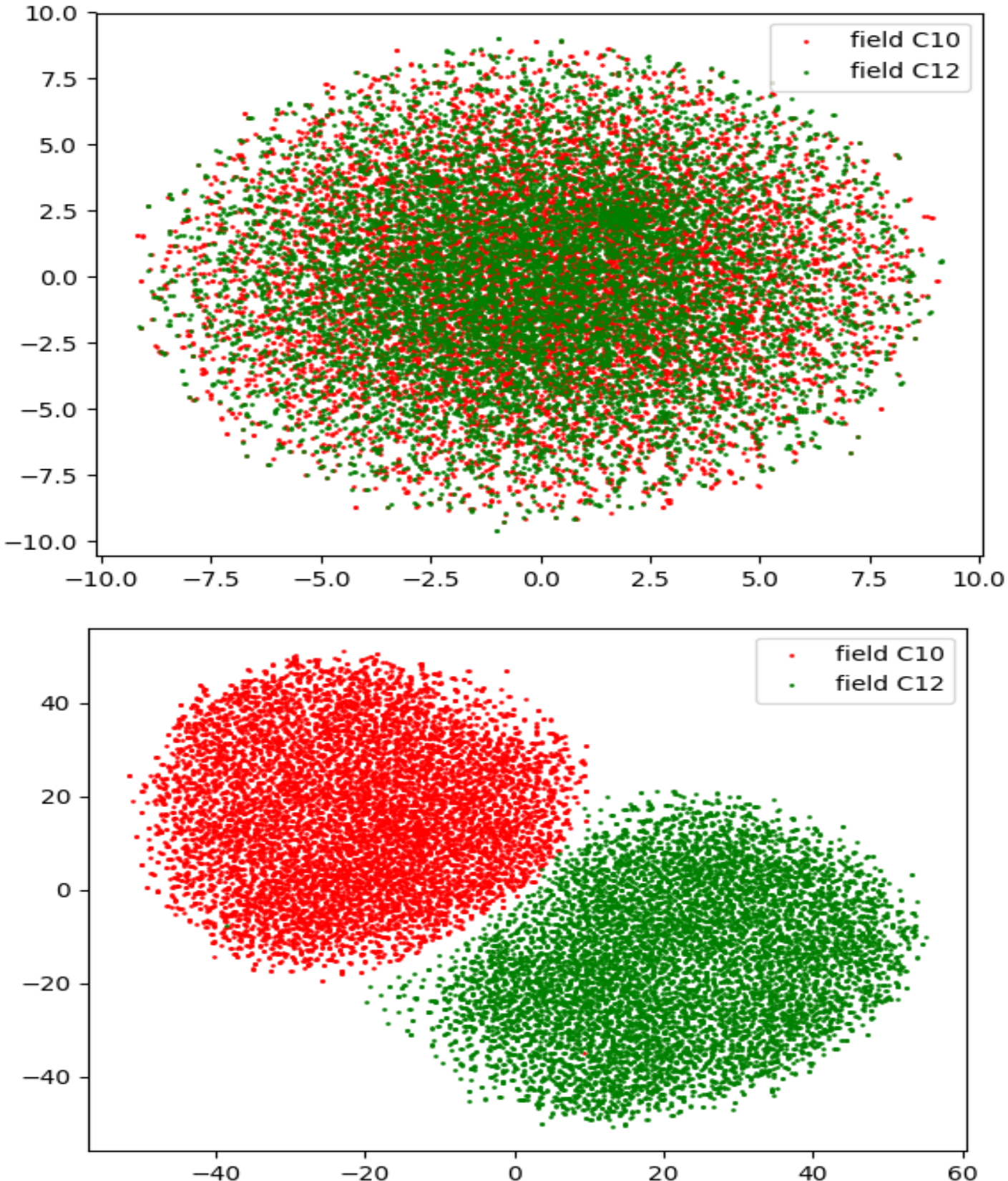}
        }
        \subfigure[Avazu]{
          \label{fig:embavazu} 
          \includegraphics[width=0.3\textwidth]{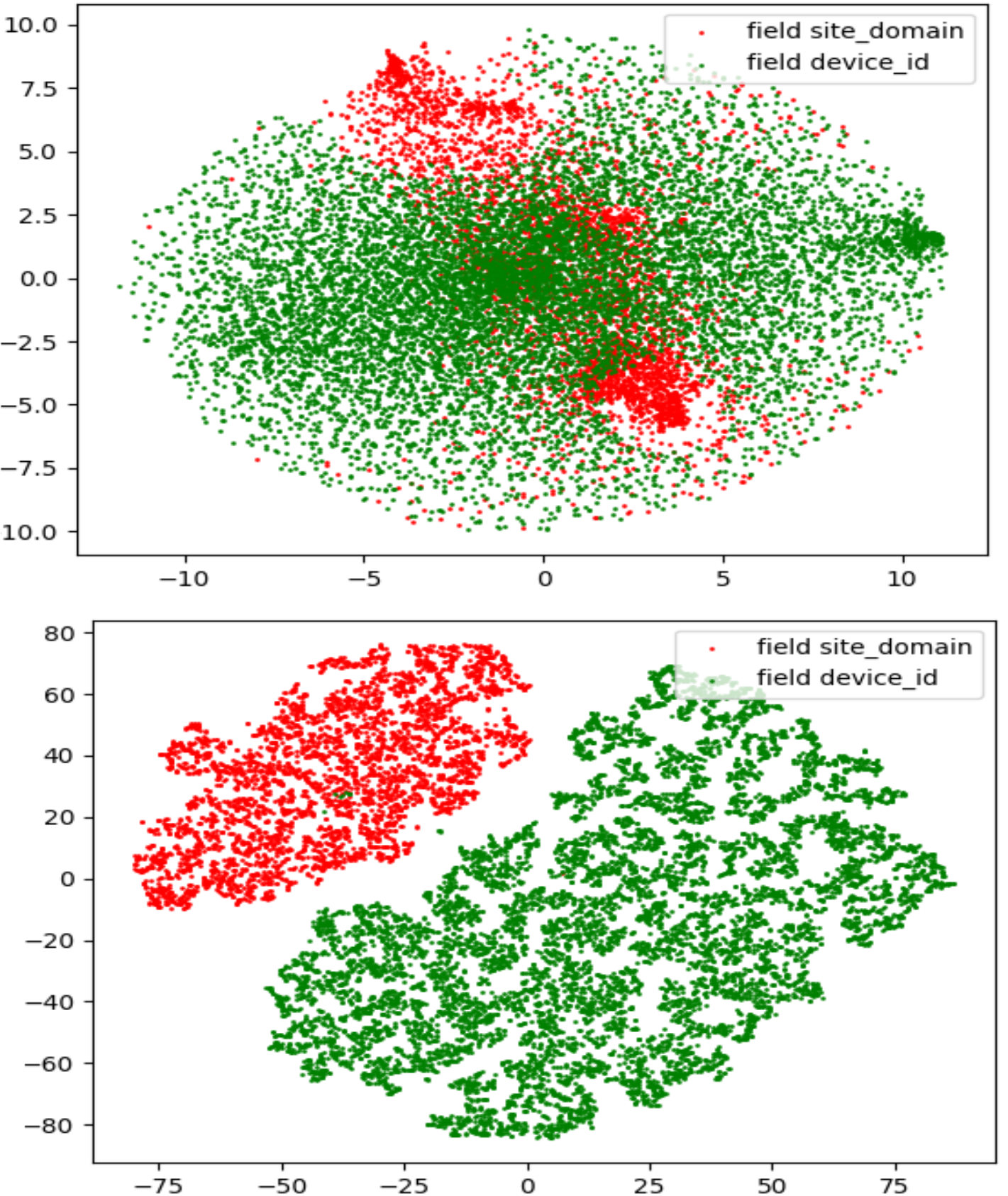}
        }
        \subfigure[Movielens]{
          \label{fig:embml} 
          \includegraphics[width=0.3\textwidth]{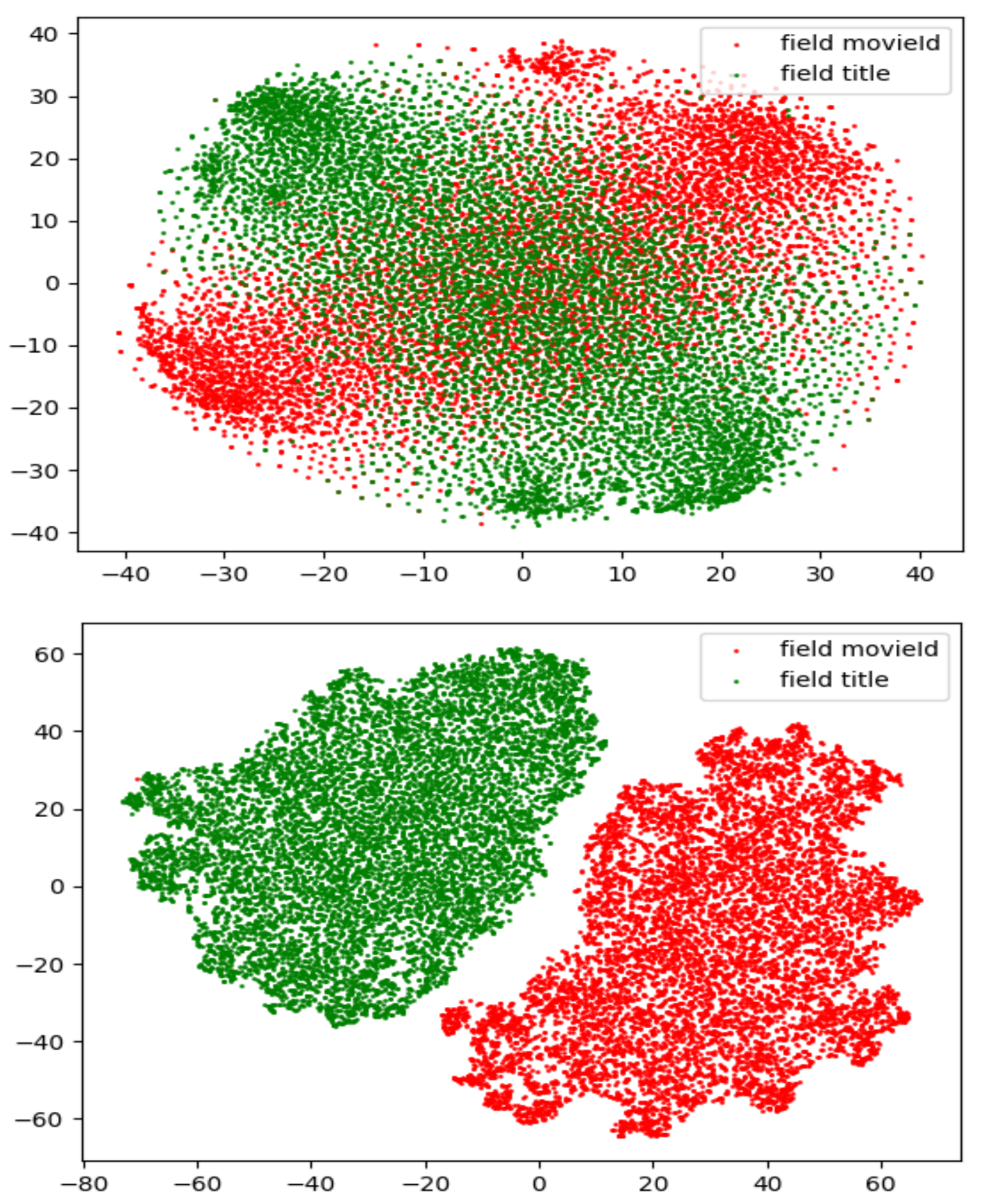}
          }
        \subfigure[Talkshow]{
        \label{fig:embluoji} 
        \includegraphics[width=0.3\textwidth]{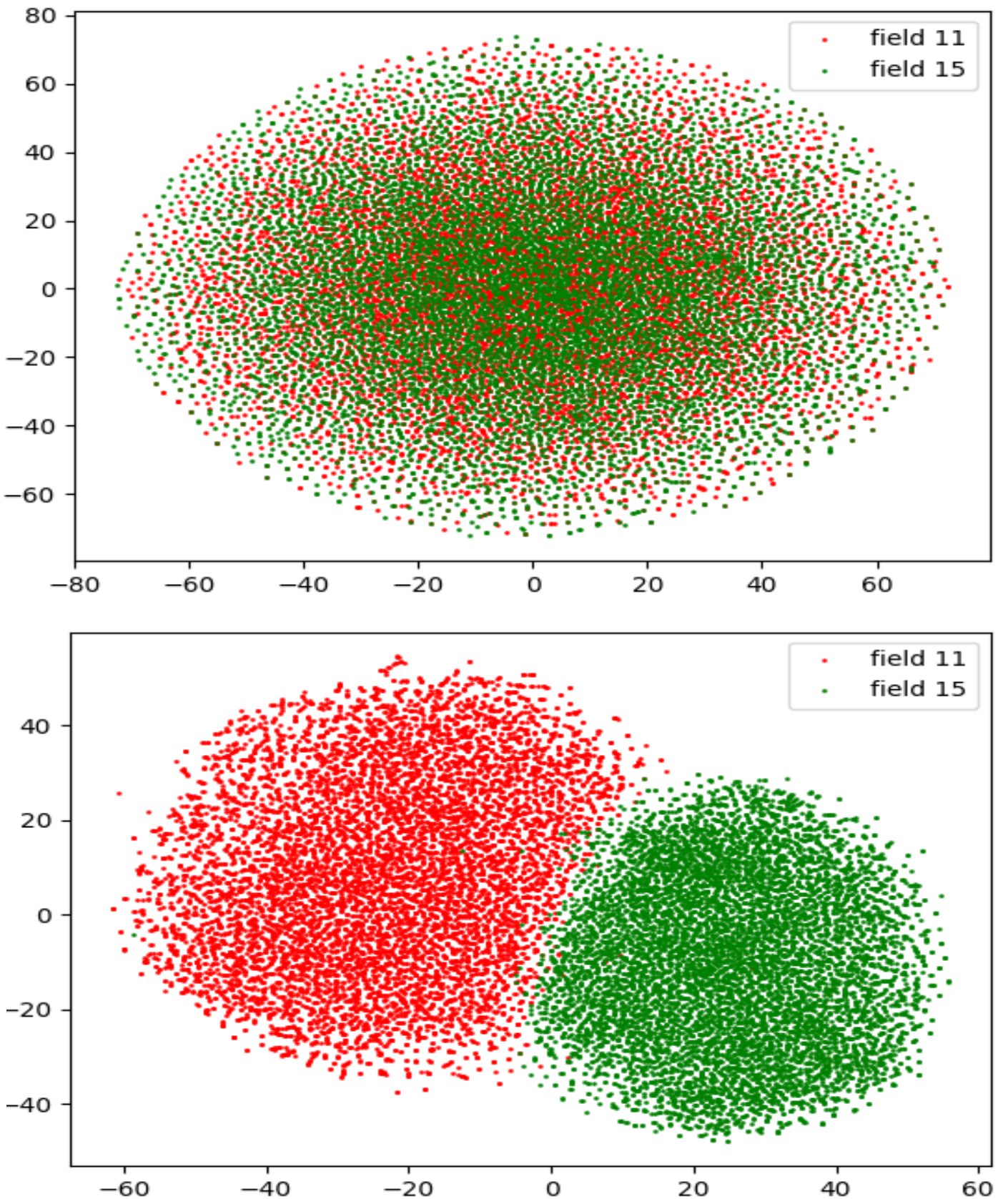}
        }
        \subfigure[Social]{
          \label{fig:embmyhug} 
          \includegraphics[width=0.3\textwidth]{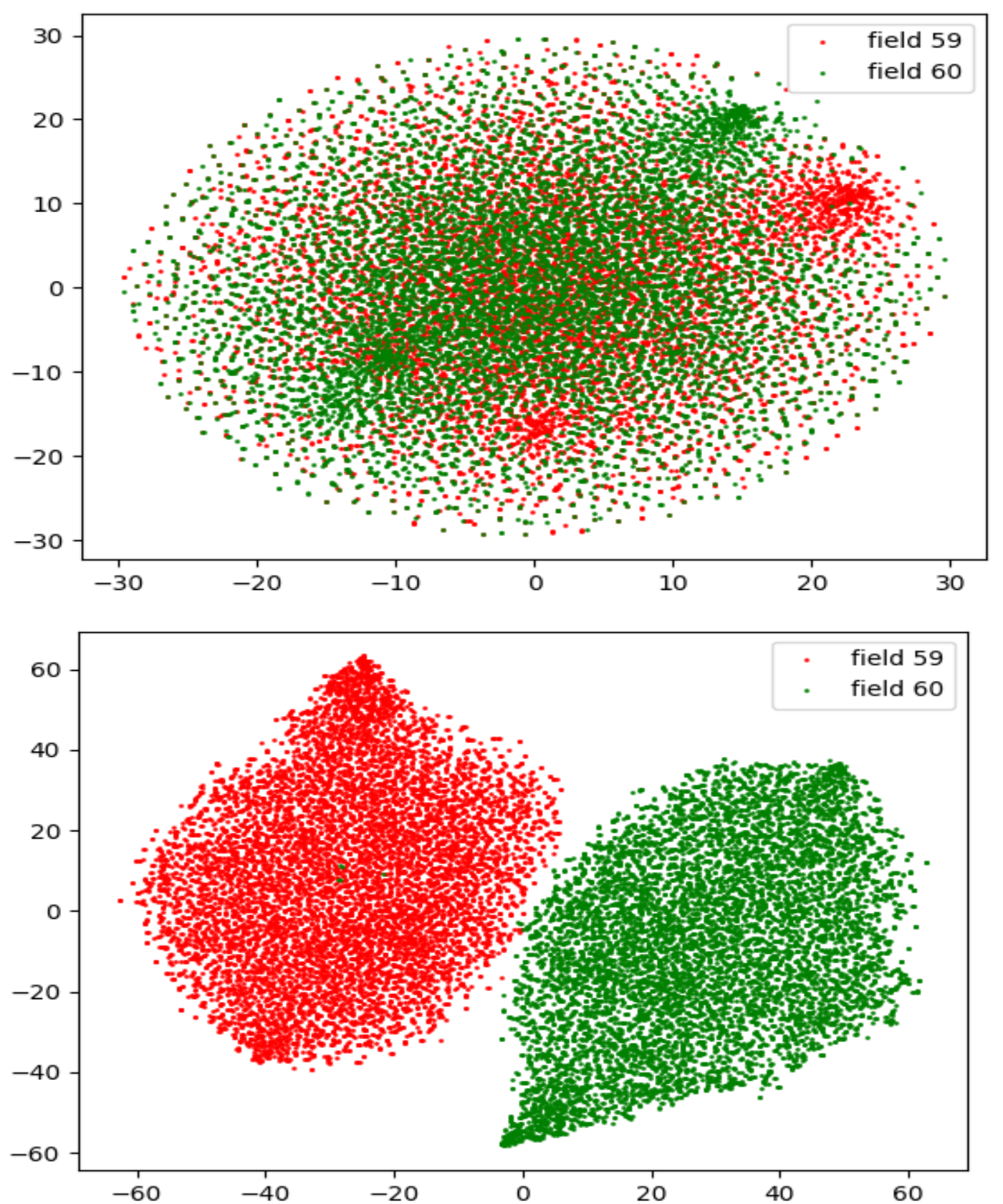}
        }
        \subfigure[Sports]{
          \label{fig:embtangdou} 
          \includegraphics[width=0.3\textwidth]{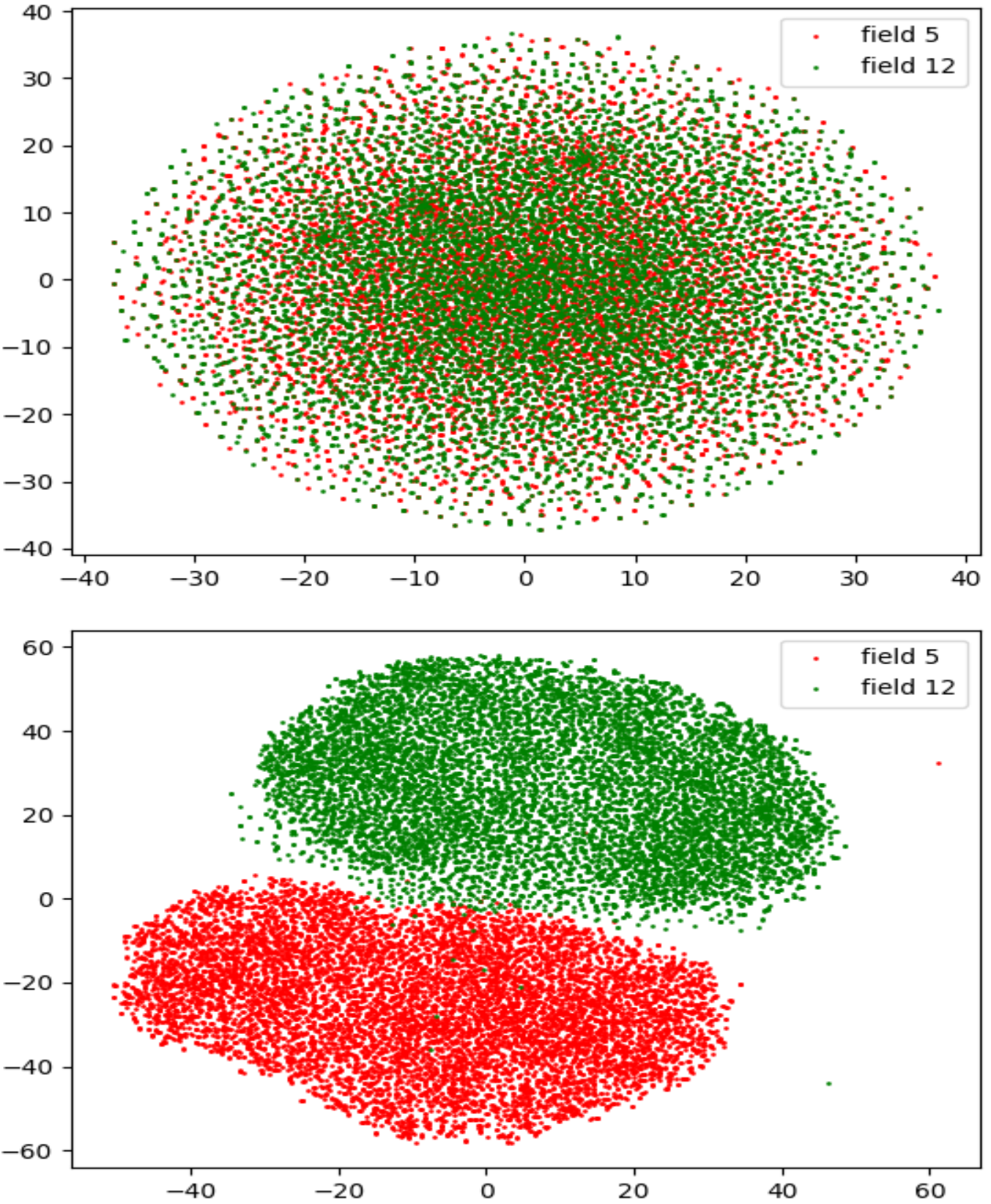}
        }
    \caption{Visualization of embeddings before (row above) and after (row below) the processing of field-wise network for different fields on six datasets (best viewed in color).
    }
    \label{fig:distributeifn}
    \vspace{-3pt}
  \end{figure*}

\vspace{-3pt}
\subsection{Performance comparison (Q2)}
\label{subsec:performance_comp}
\textbf{NON achieves state-of-the-art performance.}\quad
To further demonstrate the effectiveness of NON, we compare NON with FFM, DNN, Wide\& Deep and other three state-of-art deep models: NFM, xDeepFM, AutoInt.
Table \ref{tab:hfn} reports the test AUC and improvement on the abovementioned datasets with different methods. We have the following observations.

First and foremost, NON can always get the best results on all the datasets and achieves 0.64\%$\sim$0.99\% improvement on the test AUC compared with DNN. 
In the industrial scenario, a small improvement in offline AUC evaluation is likely to lead to significant increase in online revenue, which is also discussed in \cite{cheng2016wide,guo2017deepfm} for click-through rate prediction.
The results provide evidence to the importance of capturing intra-field information and non-linear interactions between different operations, which are the major advantages of NON.

Secondly, as compared with DNN, complex models tend to improve the performance, while sometimes also cause performance degradation. For example, NFM causes -2.97\% degradation on Talkshow dataset. 
The results indicate that increasing model capacity straightforwardly may not result in better results, and the promising results achieved by NON mainly benefits from the priority of its design paradigm.

\vspace{-4pt}
\subsection{Study of operations (Q3)}
\textbf{Most suitable operations vary from dataset to dataset.}\quad
 Because of the strong representation ability of DNN, and to avoid empty operation, we make DNN required in the across field network, while all the other operations (LR, FM, Bi-Interaction and multi-head self-attention) are optional.
The operations together with other hyper-parameters are searched simultaneously from the given space described above. 
The performance are shown in Table \ref{tab:hfn} and the most suitable operations found by the across field network for different datasets are shown in Table \ref{tab:op}. From the results, we have two observations. 
First, all of the results have LR as one of their operations, probably because linear models like LR are stable and easy to train, thus they helps model performance.
Second, operations in large datasets tend to contain complex operations like multi-head self-attention and Bi-Interaction.
While more and complex operations means larger capacity and more expressive of the model, they are prone to over-fitting and other problems. Small datasets prefer less and light operations, while large datasets require more and heavy operations to capture the rich information in big data.

To further verify the observations above, we conduct more experiments by fixing the operations in the across field network of NON. The experimental setting is same to above (Section \ref{subsec:performance_comp}) except we fix the combination of operations\footnote{The total number of combinations is 7: optional operations are \{LR, Bi-Interaction, Attention\} and empty is not taken into consideration.}.
The results are shown in Figure \ref{fig:opcombines}.
From the results, we can see that: 1) the performance changes when operations changing, and the gap between maximum and minimum test AUC ranges from 0.1\% to 0.9\% for different datasets;
2) no combination of operations can always achieve the best performance on all the datasets, which indicate the necessity to select operations in the across field network data-drivenly; 3) small datasets have better performance with less and light operations, while large datasets prefer more and complex operations.
The above observations are consistent with Table \ref{tab:op}, and
we believe this empirical study helps users to choose appropriate operations for their own tasks.

\begin{table}[!ht]
  \caption{Most suitable operations found by the across field network for different datasets.}
  \label{tab:op}
  \centering
  \renewcommand\tabcolsep{12pt}
  \begin{tabular}{lr}
    \toprule 
    Dataset      & Operations \\ 
    \midrule
    Criteo        & Attention \& DNN \& LR\\
    Avazu       & Bi-Interaction \& Attention \& DNN \& LR\\
    Movielen   & Attention \& DNN \& LR\\
    Talkshow  & DNN \& LR\\
    Social       & DNN \& LR\\
    Sports       & DNN \& LR\\
    \bottomrule 
  \end{tabular}
\end{table}

\subsection{Study of field-wise network (Q4)}
\textbf{Embeddings within each field lie closer to each other after field-wise network.}\quad
We visualize and compare the embeddings before and after the processing of field-wise network. For each dataset in Table \ref{tab:expdata}, we randomly select two categorical fields and create 2D plots using t-SNE~\cite{maaten2008visualizing}. The results are shown in Figure \ref{fig:distributeifn}, where rows above and bottom represent before and after processing respectively.
It is easy to see that after the processing of field-wise network, features within each field lie closer to each other, while features belonging to different fields are easily distinguished (the second row of Figure \ref{fig:distributeifn}).
We further calculate the micro averaged cosine distance of features over all datasets. For each field on every dataset, we first calculate the sum distance of all pairs of features, and then averaged over all the fields.
As shown in Table \ref{ta:cossim}, the results indicates field-wise network can make the cosine distance up to two order larger, which means the similarity of features within each field are captured by NON effectively.

\begin{table}[!htb]
	\caption{Micro averaged cosine distance over all fields for embeddings before and after the processing of field-wise network.}
	\label{ta:cossim}
	\centering
	\renewcommand\tabcolsep{20pt}
	\begin{tabular}{lrr}
		\toprule
		Dataset       & Before           & After \\ 
		\midrule
		Criteo         & 2.25e-5         & 2.00e-4  \\
		Avazu         & 1.43e-3         & 1.73e-2  \\
		Movielens   & 3.17e-4         & 2.95e-3  \\
		Talkshow    & 1.98e-4        & 3.14e-3 \\
    Social         & 5.41e-6         & 8.10e-4   \\
    Sports        & 1.87e-5 
    & 2.56e-4
    \\
		\bottomrule 
	\end{tabular}
\end{table}

\section{Conclusion}\label{sec:conclusion}
Tabular data classification accuracy is vital to may real-world applications and the intra-field information has been overlooked by existing deep methods.
In this paper, we propose Network On Network (NON) to provide our customers with better tabular data classification service.
NON consists of three parts to take full advantage of the intra-field information and non-linear interactions, including field-wise network at the bottom to capture intra-field information, across field network in the middle to choose appropriate operations data-drivenly, and the operation fusion network on the top to fuse outputs of the chosen operations deeply.
Experimental results on six real-world datasets show that NON can outperform several state-of-the-art models significantly and consistently, which also demonstrate the superiority of its design paradigm.
Moreover, both qualitative (i.e. visualization of features in the embedding space) and quantitative (i.e. average cosine distance) analysis show NON can capture intra-field information effectively. 

\begin{acks}
We would like to thank Guangchuan Shi, Chenggen Sun and Yingxiang Jiao for their help in the deployment of NON.
We would also like to thank Huan Zhao for the constructive suggestions and the anonymous reviewers for their insightful comments.
Qiang Yang thanks the support of Hong Kong CERG grants 16209715 and 16244616.
\end{acks}

\vfill\eject 
\bibliographystyle{ACM-Reference-Format}
\bibliography{ref}

\end{document}